\documentclass[letterpaper, 10 pt, conference]{IEEEtran}
\usepackage{etex}
\IEEEoverridecommandlockouts  

\usepackage{multicol}
\usepackage{amsthm}
\usepackage[bookmarks=true]{hyperref}

\usepackage{comment}
\usepackage[binary-units]{siunitx}
\usepackage{relsize}
\usepackage{ifthen}
\usepackage[colorinlistoftodos]{todonotes}

\usepackage[vlined,ruled,linesnumbered]{algorithm2e}
\usepackage{graphics} 
\usepackage{rotating}
\usepackage{color}
\usepackage{enumerate}
\usepackage[T1]{fontenc}
\usepackage{psfrag}
\usepackage{epsfig} 
\usepackage{booktabs}
\usepackage{graphicx,url}
\usepackage{multirow}
\usepackage{array}
\usepackage{latexsym}
\usepackage{amsfonts}
\usepackage{amsmath}
\usepackage{amssymb}
\usepackage{xstring}
\usepackage[noend]{algorithmic}
\usepackage{multirow}
\usepackage{xcolor}
\usepackage{prettyref}
\usepackage{flexisym}
\usepackage{bigdelim}
\usepackage{breqn} 
\usepackage{listings}
\let\labelindent\relax
\usepackage{enumitem}
\usepackage{xspace}
\usepackage{bm}
\graphicspath{{./figures/}}
\usepackage{tikz}
\usetikzlibrary{matrix,calc}

\usepackage{mdwlist}

\let\itemize\undefined
\makecompactlist{itemize}{stditemize}

\newcommand{\cH}{\mathcal{H}}

\newcommand{\eps}{\epsilon}

\newcommand{\nameProblemLong}{multi-model registration\xspace}
\newcommand{\nrModels}{M}

\usepackage{amsfonts,amssymb,amsmath}
\usepackage{thmtools} 
\usepackage{thm-restate}

\newtheoremstyle{mystyle}
  {}
  {}
  {\itshape}
  {}
  {\bfseries}
  {.}
  { }
  {\thmname{#1}\thmnumber{ #2}\thmnote{ (#3)}}
\theoremstyle{mystyle}

\newrefformat{prob}{Problem\,\ref{#1}}
\newrefformat{def}{Definition\,\ref{#1}}
\newrefformat{sec}{Section\,\ref{#1}}
\newrefformat{sub}{Section\,\ref{#1}}
\newrefformat{prop}{Proposition\,\ref{#1}}
\newrefformat{app}{Appendix\,\ref{#1}}
\newrefformat{alg}{Algorithm\,\ref{#1}}
\newrefformat{cor}{Corollary\,\ref{#1}}
\newrefformat{thm}{Theorem\,\ref{#1}}
\newrefformat{lem}{Lemma\,\ref{#1}}
\newrefformat{fig}{Fig.\,\ref{#1}}
\newrefformat{tab}{Table\,\ref{#1}}

\newtheorem{theorem}{Theorem}

\newtheorem{definition}[theorem]{Definition}

\newtheorem{remark}[theorem]{Remark}

\newcommand{\bdmath}{\begin{dmath}}
\newcommand{\edmath}{\end{dmath}}
\newcommand{\beq}{\begin{equation}}
\newcommand{\eeq}{\end{equation}}
\newcommand{\bdm}{\begin{displaymath}}
\newcommand{\edm}{\end{displaymath}}
\newcommand{\bea}{\begin{eqnarray}}
\newcommand{\eea}{\end{eqnarray}}
\newcommand{\beal}{\beq \begin{array}{ll}}
\newcommand{\eeal}{\end{array} \eeq}
\newcommand{\beas}{\begin{eqnarray*}}
\newcommand{\eeas}{\end{eqnarray*}}
\newcommand{\ba}{\begin{array}}
\newcommand{\ea}{\end{array}}
\newcommand{\bit}{\begin{itemize}}
\newcommand{\eit}{\end{itemize}}
\newcommand{\ben}{\begin{enumerate}}
\newcommand{\een}{\end{enumerate}}

\newcommand{\calH}{{\cal H}}

\newcommand{\eg}{\emph{e.g.,}\xspace}
\newcommand{\ie}{\emph{i.e.,}\xspace}
\newcommand{\myParagraph}[1]{{\bf #1.}\xspace}

\newcommand{\M}[1]{{\bm #1}} 
\renewcommand{\boldsymbol}[1]{{\bm #1}}

\newcommand{\hide}[1]{}

\newcommand{\hiddenText}{{\color{gray} hidden text.}}
\newcommand{\hideWithText}[1]{\hiddenText}

\DeclareMathOperator*{\argmax}{arg\,max}

\newcommand{\normsq}[2]{\left\|#1\right\|^2_{#2}}
\newcommand{\norm}[1]{\left\| #1 \right\|}

\newcommand{\trace}[1]{\mathrm{tr}\left(#1\right)}

\newcommand{\e}{{\mathrm e}}

\newcommand{\zero}{{\mathbf 0}}
\newcommand{\eye}{{\mathbf I}}

\newcommand{\Real}[1]{ { {\mathbb R}^{#1} } }

\newcommand{\at}[1]{^{(#1)}}

\newcommand{\SOthree}{\ensuremath{\mathrm{SO}(3)}\xspace}

\newcommand{\MR}{\M{R}}

\newcommand{\va}{\boldsymbol{a}} 
 
\newcommand{\vb}{\boldsymbol{b}}

\newcommand{\vf}{\boldsymbol{f}}

\newcommand{\vo}{\boldsymbol{o}}

\newcommand{\vt}{\boldsymbol{t}}
\newcommand{\vxx}{\boldsymbol{x}} 
\newcommand{\vy}{\boldsymbol{y}}

\newcommand{\vtheta}{\boldsymbol{\theta}}

\newcommand{\vepsilon}{\boldsymbol{\epsilon}}

\newcommand{\blue}[1]{{\color{blue}#1}}

\newcommand{\linkToPdf}[1]{\href{#1}{\blue{(pdf)}}}
\newcommand{\linkToPpt}[1]{\href{#1}{\blue{(ppt)}}}
\newcommand{\linkToCode}[1]{\href{#1}{\blue{(code)}}}
\newcommand{\linkToWeb}[1]{\href{#1}{\blue{(web)}}}
\newcommand{\linkToVideo}[1]{\href{#1}{\blue{(video)}}}
\newcommand{\linkToMedia}[1]{\href{#1}{\blue{(media)}}}
\newcommand{\award}[1]{\xspace} 

\usepackage{float}
\usepackage{graphicx}
\usepackage{epstopdf}
\usepackage[font=small]{caption}
\usepackage{epsfig}
\usepackage{cleveref}
\usepackage{subcaption}
\theoremstyle{definition}
\newcommand{\myparagraph}[1]{{\bf #1.}}

\usepackage{xr}

\title{\huge{Multi-Model 3D Registration: Finding Multiple \\ Moving Objects in Cluttered Point Clouds}} %

\author{David Jin, Sushrut Karmalkar, Harry Zhang, Luca Carlone \vspace{-5mm}
\thanks{D.\,Jin, H.\,Zhang, and L.\,Carlone are with the Laboratory for 
Information \& Decision Systems, Massachusetts Institute of Technology, Cambridge, MA, USA, 
{\sf \{jindavid,harryz,lcarlone\}@mit.edu}
}
\thanks{S.\,Karmalkar is with the Department of Computer Science at the University of Wisconsin at Madison, Madison, WI, USA, 
{\sf s.sushrut@gmail.com}
}
}

\begin{document}

\maketitle

\begin{tikzpicture}[overlay, remember picture]
\path (current page.north east) ++(-2.6,-0.2) node[below left] {
This paper has been accepted for publication in the IEEE International Conference on
Robotics and Automation.
};
\end{tikzpicture}
\begin{tikzpicture}[overlay, remember picture]
\path (current page.north east) ++(-5.3,-0.6) node[below left] {
Please cite the paper as: D. Jin, S. Karmalkar, H. Zhang, and L. Carlone,
};
\end{tikzpicture}
\begin{tikzpicture}[overlay, remember picture]
\path (current page.north east) ++(-3.9,-1) node[below left] {
``Multi-Model 3D Registration: Finding Multiple Moving Objects in Cluttered Point Clouds’’,
};
\end{tikzpicture}
\begin{tikzpicture}[overlay, remember picture]
\path (current page.north east) ++(-5.3,-1.4) node[below left] {
 \emph{IEEE International Conference on
Robotics and Automation (ICRA)}, 2024.
};
\end{tikzpicture}


\begin{abstract}
We investigate a variation of the 3D registration problem, named \emph{multi-model 3D registration}. In the \nameProblemLong problem, we are given two point clouds picturing a set of objects at different poses (and possibly including points belonging to the background) and we want to simultaneously reconstruct how all objects moved between the two point clouds.
 This setup generalizes standard 3D registration where one wants to reconstruct a single pose, \eg the motion of the sensor picturing a static scene. Moreover, it provides a mathematically grounded formulation for relevant robotics applications, \eg where a depth sensor onboard a robot perceives a dynamic scene and has the goal of estimating its own motion (from the static portion of the scene) while simultaneously recovering the motion of all dynamic objects. 
We assume a correspondence-based setup where we have putative matches between the two point clouds and consider the practical case where these correspondences are plagued with outliers.
We then propose a simple approach based on Expectation-Maximization (EM) and 
establish theoretical conditions under which the EM approach converges to 
the ground truth. We evaluate the approach in simulated and real datasets ranging from table-top 
scenes to  self-driving scenarios and demonstrate its effectiveness 
when combined with state-of-the-art 
scene flow methods 
to establish dense correspondences.  
\end{abstract}
\vspace{-3mm}
\section{Introduction}
\label{sec:intro}

3D registration is a foundational problem in robotics and computer vision and arises 
in several applications, including motion estimation and 
3D reconstruction~\cite{Henry12ijrr-rgbdMapping,Blais95pami-registration,Choi15cvpr-robustReconstruction},
object pose estimation~\cite{Drost10cvpr,Wong17iros-segicp,Zeng17icra-amazonChallenge}, 
and medical imaging~\cite{Audette00mia-surveyMedical,Tam13tvcg-registrationSurvey};
rotation-only variations of the problem also arise in
panorama stitching~\cite{Bazin14eccv-robustRelRot} and satellite attitude determination~\cite{wahba1965siam-wahbaProblem}.  

\myParagraph{3D Registration}
In its simplest form, 3D registration looks for the rotation $\MR\in\SOthree$ and translation $\vt\in\Real{3}$ that align two sets of points $\{\va_i\}_{i=1}^n$ and $\{\vb_i\}_{i=1}^n$. If the correspondences between the two sets of points are known, \ie we know that point $\vb_i$ in the second point cloud corresponds to point $\va_i$ in the first point cloud after a rigid transformation $(\MR,\vt)$ is applied, then the problem can be formulated as a nonlinear least squares problem and solved in closed form~\cite{Arun87pami,Horn87josa}. More formally, if we assume the following generative model 
\beq
\label{eq:measModel1}
\vb_i = \MR \va_i + \vt + \vepsilon, \quad i=1,\ldots,n
\eeq
where $\vepsilon$ is a noise term distributed according to an isotropic Gaussian, then a maximum likelihood estimate for $(\MR, \vt)$ can be computed by solving the following nonlinear least squares:
\beq
\label{eq:estimator1} 
\min_{\MR\in\SOthree, \vt\in\Real{3}} \textstyle\sum_{i=1}^n \normsq{ \vb_i - \MR \va_i - \vt }{}
\eeq
which admits a well-known closed-form solution via singular value decomposition (SVD)~\cite{Arun87pami,Horn87josa}.

\begin{figure}
    \centering
    \includegraphics[width=\columnwidth]{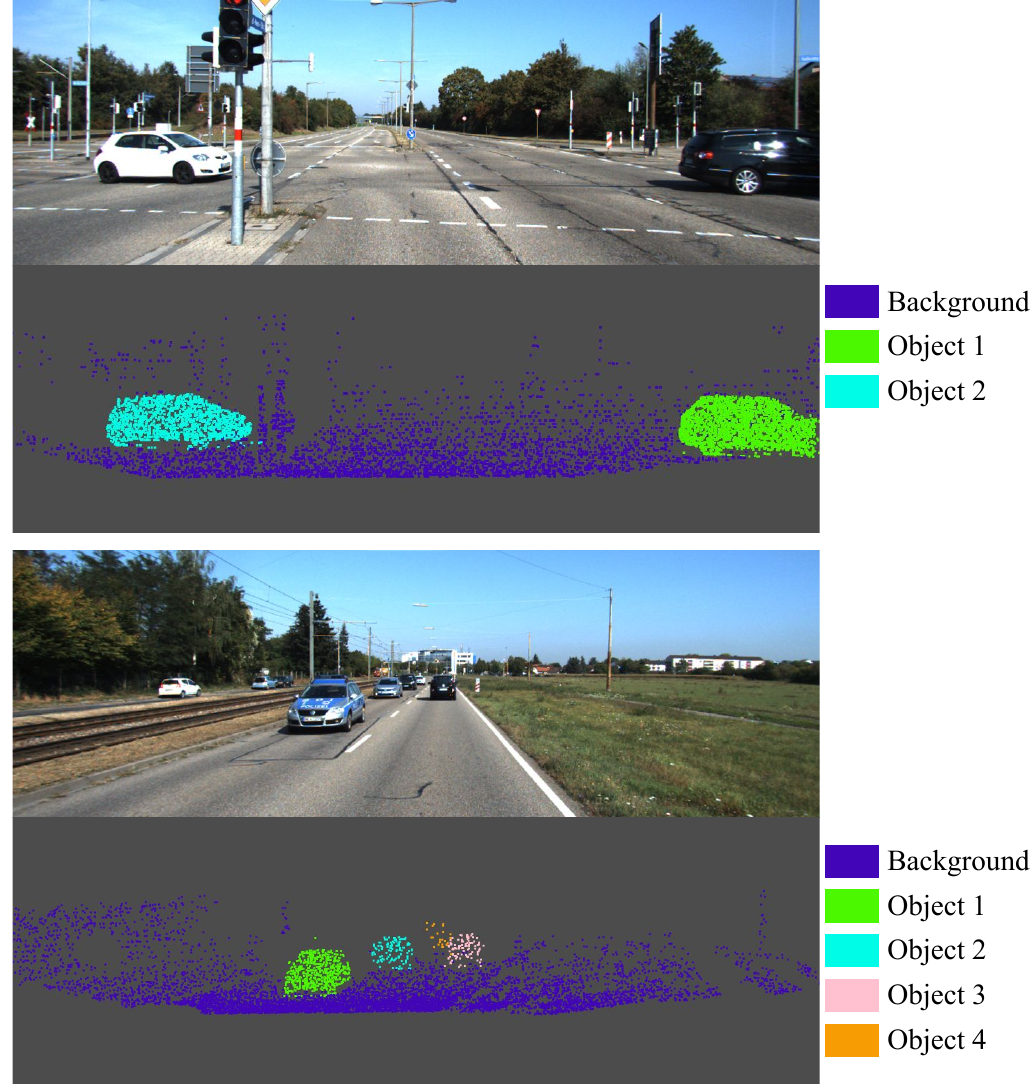}
    \vspace{-4mm}
    \caption{We propose an Expectation-Maximization approach for \emph{multi-model 3D registration}, which aims to recover the motion of all objects (and background) in a scene  from point cloud observations. The figure reports two results produced by our approach on the KITTI dataset.
    Note that the two cars on the left of the bottom figure are stationary, hence they are correctly deemed to be part of the background.
    \hspace{-3mm}
    \label{fig:coverfigure}\vspace{-4mm}}
\end{figure}

\myParagraph{Robust 3D Registration} In practical problems, the measurements contain spurious correspondences. For instance, if the two point clouds represent two RGB-D scans at consecutive time stamps and we are trying to estimate the motion of the sensor between scans, we might attempt to establish correspondences $(\vb_i,\va_i)$, $i=1,\ldots,n$, using descriptor matching~\cite{Yang20tro-teaser}, optical flow~\cite{Barron94}, or scene flow~\cite{Vedula05pami-sceneFlow}.
As a result, some of the point pairs $(\vb_i,\va_i)$ may be well-approximated by the measurement model~\eqref{eq:measModel1}, while others are \emph{outliers} and largely deviate from~\eqref{eq:measModel1}, either because the pairs of points are incorrectly associated by the algorithm that establishes the correspondences, or because they do not lie on a static portion of the scene.
This has motivated a large body of work on \emph{robust 3D registration}, which focuses on estimating $(\MR,\vt)$ in the face of outliers. In this case, the measurement model becomes:
\vspace{-2mm}
\beq
\label{eq:measModel2}
\vb_i = \theta_i (\MR \va_i + \vt) + (1-\theta_i) \vo + \vepsilon, \quad i=1,\ldots,n
\eeq
where the (unknown) binary variable $\theta_i\in\{0,1\}$ decides whether $\vb_i$ is a rigid transformation of $\va_i$ (if $\theta_i=1$) or is an arbitrary vector $\vo$, independent of $(\MR,\vt)$ (if $\theta_i=0$). 
A plethora of works has attacked robust registration with outliers.
While we refer the reader to Section~\ref{sec:relatedWork} and~\cite{Yang20tro-teaser} for a more extensive discussion about related work, a popular approach is to resort to M-estimation, which attempts to compute an estimate for $(\MR, \vt)$ by minimizing a robust loss function. For instance, the work~\cite{Yang20tro-teaser} considers a truncated least squares loss:

\vspace{-3mm}
\beq
\label{eq:estimator2}
\min_{ \substack{\MR\in\SOthree, \vt\in\Real{3}, \\ \theta_i \in \{0;1\}, i=1,\ldots,n}  } 
\textstyle\sum_{i=1}^n  \theta_i \normsq{ \vb_i - \MR \va_i - \vt }{} + (1-\theta_i) \bar{c}^2
\eeq
which computes a least squares estimate for measurements with small residual errors (\ie whenever $\|\vb_i - \MR \va_i - \vt\|<\bar{c}$ the optimization forces $\theta_i=1$ and the second summand disappears), while discarding measurements with large residuals (when $\theta_i=0$, the objective becomes constant and the $i$-th measurement does not contribute to the estimate). 

\myParagraph{Multi-Model 3D Registration}
The robust registration problem~\eqref{eq:estimator2} looks for a single pose that explains the majority of  correspondences, while disregarding the others as outliers. 
In this paper, we ask: can we instead find further patterns in the outliers? 
or, in other words, can we \emph{simultaneously recover the motion of all 
objects present in the point clouds}?
More formally, we assume the following generative model: 
\beq
\label{eq:measModel3}
\vb_i = \textstyle\sum_{j=1}^\nrModels \theta_{i,j} (\MR_j \va_i + \vt_j) + \theta_{i,0} \; \vo + \vepsilon, \quad i=1,\ldots,n
\eeq
where for each measurement $i$, the vector $\vtheta_i = [\theta_{i,0} \; \theta_{i,1} \; \ldots \theta_{i,\nrModels}] \in \{0;1\}^{\nrModels+1}$  is an unknown binary vector with a single entry equal to 1, $\nrModels$ it the number of objects (unknown a priori), and $(\MR_j,\vt_j)$ is the motion of the $j$-th object, for $j=1,\ldots,\nrModels$. In words, each point $\vb_i$ in~\eqref{eq:measModel3} is either generated by an object $j$ (if $\theta_{i,j}=1$ for a $j\in\{1,\ldots,\nrModels\}$) or is an outlier (if $\theta_{i,0}=1$). 
Clearly, when $\nrModels=1$, eq.~\eqref{eq:measModel3} falls back to the  robust registration setup in~\eqref{eq:measModel2}. 

\myParagraph{Contribution}
We propose an approach to solve the \nameProblemLong in eq.~\eqref{eq:measModel3}. 
The approach is based on an Expectation-Maximization (EM) algorithm that computes the assignments 
 of measurements to objects (\ie the vectors $\vtheta_i$ in~\eqref{eq:measModel3}) and retrieves the pose $(\MR_j,\vt_j)$ for each object. The approach does not require prior knowledge of the number of objects $\nrModels$ and can also accommodate additional constraints (\eg that distant objects are distinct, even if they exhibit similar motion). We provide a novel theoretical analysis of the algorithm that suggests that the EM scheme converges to the ground truth as long as the initialization of the vectors $\vtheta_i$ is sufficient to capture all objects of interest. 
 We evaluate the EM scheme in simulated and real datasets ranging from table-top 
scenes to large self-driving scenarios (Fig.~\ref{fig:coverfigure}) and demonstrate its effectiveness when combined with state-of-the-art 
scene flow methods 
to establish dense correspondences.  
\section{Related work}
\label{sec:relatedWork}

\myParagraph{Robust Estimation in Robotics and Vision}
Robust estimation is an active research area in robotics and vision~\cite{Peng23CVPR-IRLS,Barik23arxiv-invexity,Carlone23fnt-estimationContracts} and has been attacked using different frameworks, including M-estimation~\cite{MacTavish15crv-robustEstimation,Black96ijcv-unification,Yang20ral-GNC,Yang22pami-certifiablePerception}, 
consensus maximization~\cite{Chin18eccv-robustFitting,Antonante21tro-outlierRobustEstimation} (typically solved using sampling-based algorithms, such as RANSAC~\cite{Fischler81}), or graph-theoretic  methods~\cite{Yang20tro-teaser,Shi21icra-robin,Parra19arXiv-practicalMaxClique,Enqvist09iccv}.
We refer the reader to~\cite{Yang20tro-teaser} for an extensive review of robust 3D registration and to~\cite{Carlone23fnt-estimationContracts,Bosse17fnt} for an overview of robust estimation across robotics and vision.

\myParagraph{List-Decodable Regression}
While standard robust estimation computes an estimate that agrees with the majority of the measurements, 
recent work in robust statistics has focused on recovering an estimate from a handful of inliers hidden among an overwhelming amount of outliers, \eg~\cite{Charikar17stoc-robustEstimationTheory,Karmalkar19neurips-ListDecodableRegression, Raghavendra20soda-ListDecodableRegressions,diakonikolas2020list,cherapanamjeri2020list}. In this regime, returning a  
single accurate hypothesis is information-theoretically impossible, 
and one has to compute a list of hypotheses to guarantee that at least one of them is accurate.
This setup, typically referred to as \emph{list-decodable} regression, 
was first studied in \cite{Karmalkar19neurips-ListDecodableRegression} and~\cite{Raghavendra20soda-ListDecodableRegressions}, which proposed and analyzed algorithms based on semidefinite relaxations. The work~\cite{Carlone23fnt-estimationContracts} observes that the algorithm in~\cite{Karmalkar19neurips-ListDecodableRegression} can be easily adapted to solve a multi-model rotation-only registration problem; however, the resulting relaxation is impractically slow (\eg 3 minutes to solve a problem with 50 measurements).

\myParagraph{Multi-Model Fitting in Computer Vision} Early work in computer vision has studied the problem of simultaneously recovering multiple models from noisy measurements. 
The corresponding literature includes clustering-based and optimization-based methods.
Clustering-based methods span a variety of techniques, including  hierarchical clustering~\cite{Magri14CVPR-tlinkage}, \cite{Roberto08ECCV-jlinkage}, kernel fitting~\cite{Chin09ICCV-statlearning}, \cite{Chin10CRPR-kerneloptim}, matrix factorization~\cite{Magri17IVC-RPA}, \cite{Mariano17CVPR-NMU} and hypergraph partitioning~\cite{Lin19AAAI-hypermultifitting}, \cite{Purkait17TPAMI-clusterhypergraph}. 
Optimization-based methods include generalizations of RANSAC to the multi-model setup, including 
Sequential RANSAC~\cite{Torr98royalp-seqransac}, Multi-RANSAC~\cite{Zuliani05ICIP-multiransac}, and RANSACOV~\cite{Magri16CVPR-ransacov}.
 Other methods such as Pearl~\cite{Isack12IJCV-pearl} and Progressive-X~\cite{Daniel19ICCV-progx} {take a step further by incorporating additional priors into the objective function.}
%
%
%

\myParagraph{Mixture of Linear Regression in Applied Mathematics} 
In the problem of learning a mixture of linear regressions, 
each measurement is generated from one of several unknown linear regression components, 
and one has to associate measurements to components and estimate the components.
This problem is known to be NP hard in general~\cite{yi2014alternating}. 
However, under certain assumptions on the underlying distribution (\eg the regressors follow a standard Gaussian distribution, or there are only two mixture components),
several approaches have successfully tackled the problem,
including algorithms based on the method of moments \cite{sedghi2016provable,li2018learning}, 
alternating-minimization \cite{yi2014alternating,YiCS16}, and 
Expectation-Maximization \cite{FariaSoromenho10,KYB19,kwon20aistats-EM-MLR}. 
Contrary to this line of work, in this paper we do not make a Gaussian assumption on the regressors, 
and we consider a 3D registration problem 
rather than a linear regression setup.

\myParagraph{Learning-based Methods for Motion Tracking} 
Learning-based methods have recently demonstrated excellent performance in 2D and 3D motion tracking.
Optical flow methods~\cite{Teed21ECCV-raft,Ilg16arxiv-flownet2} estimate the pixel displacement between two  frames, which can be used to segment moving objects from a video. 
Scene flow estimates dense 3D motion for each pixel from a pair of stereo or RGB-D frames. Other approaches, such as DRISF~\cite{Ma19CVPR-drisf} and RigidMask~\cite{Yang21CVPR-rigidmask}, divide scene flow estimation into multiple subtasks and build modular networks to solve each subtask.
RAFT-3D~\cite{Teed21CVPR-raft3d} computes the scene flow by using feature-level fusion.
CamLiRAFT~\cite{Liu22CVPR-camliflow,Liu23arxiv-camliraft} proposes a multi-stage pipeline to better fuse multi-modal information without suffering from accuracy loss due to voxelization.







\section{An Expectation-Maximization Approach to Multi-Model Registration}
\label{sec:xxx}





The Expectation-Maximization (EM) algorithm~\cite{Moon96spm} iteratively estimates parameters in statistical models given noisy data, by alternating an Expectation (E) step and a Maximization (M) step. 
In robotics, EM has been a popular approach to attack estimation problems including discrete and continuous variables~\cite{Eckart18cvpr-HGMR, Rogers10IROS-EMslam,Indelman14ICRA-posegraph,Bowman17icra}.
Here we use a variation of the EM algorithm known as the ``Classification Expectation-Maximization'' algorithm (\eg~\cite{FariaSoromenho10}), see~\Cref{alg:EM}. 
\vspace{-5mm}

\begin{algorithm}
\caption{Expectation-Maximization (EM)}
\label{alg:EM}
\SetAlgoLined
\small
\KwIn{
 Point clouds $S := \{(\va_i,\vb_i)\}_{i=1}^n$,
Initial clusters $\cH := \{H_j \subset S \mid j \in [K] \}$, 
Distance threshold $\tau$, 
Number of iterations $T$,
Minimum cluster size $m_{\min}$}
\KwOut{$H_j, \MR_j^{(r)}, \vt_j^{(r)}, \forall j \in [K]$}
\smallskip

\For{$r \in [T]$}{

{\color{gray} \% Compute a pose, weight, and variance for each cluster}
\For{$j \in [K]$}{

$(\MR_j^{(r)}, \vt_j^{(r)}) := \text{Horn}(H_j)$. \label{line:procrustes}
$\pi_j^{(r)} := |H_j|/n$. \label{line:weight} \\
$E_{j} := \{ \vb - \MR_j^{(r)} \va - \vt_j^{(r)} \mid (\va, \vb) \in H_j \}$\\
$\hat\sigma_j^{(r)} := \sqrt{ \frac{1}{3} \trace{ \text{cov}(E_j) }}$ \label{line:variance} 
}
\smallskip
\textbf{E-step:} {\color{gray} \% Compute weighted likelihood:} \\
\smallskip
\For{$j \in [K]$ and $i \in [n]$}{
$ W_{i,j; \tau}^{(r)}$ := eq.~\eqref{eq:likelihood}
}
\smallskip
{\color{gray} \% Remove small clusters $H_j$ from $\cH$}

\For{$j \in [K]$ \label{line:remove}}{
\If{$|H_j| < m_{\min}$} {
remove cluster $j$ from $\calH, \pi_j^{(r)}, \hat\sigma_j^{(r)}, W_{i,j; \tau}^{(r)}$
$K := K - 1$ 
}
}
\smallskip
\textbf{M-Step:} {\color{gray} \% Regenerate clusters according to likelihoods} \\
\smallskip

\For{$i \in [n]$}{
\If{$j^\star = \argmax_{j \in [k]} W^{(r)}_{i, j;\tau}$}{
add $(\va_i, \vb_i)$ to cluster $H_{j^\star}$
}
}

}
\end{algorithm}
\vspace{-4mm}

We start by observing that finding the associations $\vtheta_i$ can be equivalently thought of as a \emph{clustering} problem, where we try to cluster together measurements corresponding to the same object. 
We will refer to our clusters with $H_j \subset S$, where $S$ is the given set of correspondences $\{ (\va_i, \vb_i)\}_{i=1}^n$ and $H_j$ indicates the correspondences (putatively) associated with object $j$. 
Note that this interpretation is consistent with~\eqref{eq:measModel3}, and 
by definition $H_j := \{ (\va_i, \vb_i) \in S \mid \theta_{i,j} = 1 \}$.
 Accordingly, in \Cref{alg:EM}, rather than updating the indicator vectors $\vtheta_i$, 
 we update the clusters $H_j$ for all objects $j$, at each iteration.

\myparagraph{Initialization}
The algorithm takes as input, an initial guess for the clusters 
$\cH := \{H_j \subset S \mid j \in [K]\}$ of the correspondences
$S$, 
where for each object $j \in [K]$, $H_j$ is the set of correspondences associated to $j$. In the next section, we provide conditions on the initialization under which the EM algorithm converges to the ground truth.


\myparagraph{EM Algorithm} Each iteration of \Cref{alg:EM} performs an E-step and M-step. 
At each iteration $r$, the algorithm first computes a transform $(\MR_j^{(r)}, \vt_j^{(r)})$ for each cluster (line~\ref{line:procrustes}); this is done using Horn's method \cite{Horn87josa} given the measurements in that cluster.
The algorithm also computes a weight $\pi_j^{(r)}$ (quantifying the relative size of  cluster $j$) and an intra-cluster variance $\hat\sigma_j^{(r)}$ for each cluster (lines~\ref{line:weight}-\ref{line:variance}). 
Then, the E-step estimates the posterior probability that the data point 
$(\va_i, \vb_i)$ belongs to the cluster $j$ according to the weighted likelihood:

\vspace{-5mm}
 \begin{equation}
W^{(r)}_{i,j;\tau} :=\frac{\pi_j^{(r)} \phi^{(r)}(\vb_i | \va_i)}{\sum_{j=1}^{k} \pi_j^{(r)} \phi^{(r)}(\vb_i | \va_i) } \cdot \mathbf{1}(d_{\text{cluster}}(H_j, (\va_i, \vb_i)) < \tau).
\label{eq:likelihood}
\end{equation}
Here $\phi_j^{(r)}(\vb_i|\va_i)$ denotes the likelihood of $\vb_i - \MR_j^{(r)} \va_i - \vt_j^{(r)}$ with respect to the  multivariate Gaussian density with mean $\zero$ and covariance $\hat\sigma_j^2 \eye_3$. 
The first term of the likelihood essentially quantifies how well the transformation $(\MR_j^{(r)},\vt_j^{(r)})$ agrees with the correspondence $(\vb_i,\va_i)$; the weighted likelihood also accounts for the cluster size (\ie the weight $\pi_j$). 
The second term $\mathbf{1}(d_{\text{cluster}}(H_j, (\va_i, \vb_i)) < \tau)$ assigns zero likelihood to points that are far away (farther than a distance $\tau$) from cluster $j$,  
where  
$ d_{\text{cluster}}(H_j, (\va_i, \vb_i)) := \min_{\va' \in H_j} \norm{\va' - \va_i}$. 
This term avoids to cluster together objects that might have the same motion, but are far away from each other. 

The M-step 
updates the assignment of samples to the clusters by assigning each $(\vb_i,\va_i)$ to the cluster $H_j$ maximizing $W^{(r)}_{i,j;\tau}$. 
This particular variation of the M-step is called the ``Classification M-step'', see, \eg~\cite{FariaSoromenho10}. Before executing the M-step, the algorithm removes overly small clusters (line~\ref{line:remove}). 

We remark that \Cref{alg:EM} is almost parameter free, and only requires setting the distance $\tau$ beyond which we consider two objects to be distinct, the minimum ``size'' $m_{\min}$ of what we would consider an object, and the number of iterations $T$. In particular, the weighted likelihood only depends on $\tau$ and does not require setting a noise bound, \eg as done in RANSAC.
We also remark that the number of clusters $K$ is estimated during the iterations, and ideally will converge to the true number of objects $M$, see eq.~\eqref{eq:measModel3}.

In the following section, we derive conditions under which 
\Cref{alg:EM} converges to the ground truth clusters.

\section{Theoretical Analysis}
\label{sec:theory}

In this section, we sketch a proof demonstrating that \Cref{alg:EM} recovers the ground truth clusters 
under suitable condition on the initial clusters.
In the following, we say that a set of points $P$ is 
\emph{$\tau$-connected} if between any pair of points $\vxx, \vy \in P$ there is a sequence of points in $P$ such that each pair of consecutive points in the sequence are at most at distance $\tau$ from each other. 
Then we say that a set of correspondences $S := \{(\va_i, \vb_i) \}_{i=1}^n$ is $\tau$-connected if $\{\va_i\}_{i=1}^n$ is $\tau$-connected.

We make the following assumptions on the ground truth. 

\begin{definition}[Ground Truth]\label{set:problem}
We are given a set of correspondences $S := \{(\va_i, \vb_i) \}_{i=1}^n$, 
which can be partitioned into $M$ $\tau$-connected parts $G_1, \dots, G_M$ such that
there are functions $g_1, \dots, g_M$ of the form $g_j = \MR_j \va + \vt_j$ where $(\MR_j, \vt_j)$ is a rigid transformation, satisfying:
\begin{enumerate}
\item \textbf{Uniform Bounded Noise:} 
For all $j \in [M]$ and $(\va, \vb) \in G_j$, 
$g_j(\va) + \vepsilon = \vb$, 
where $\vepsilon$ is drawn from the uniform distribution over $ [-\sigma, \sigma]^3$. \label{item:GT}
\item \textbf{Object Separation:} 
For all distinct $i, j \in [M]$, 
$\min_{(\va,\vb) \in G_i, (\va',\vb') \in G_j} \norm{\va - \va'} > \tau$. \label{item:ObjSep}
\item \textbf{Bounded Point Cloud:} 
For all $i \in [n]$, $\norm{\va_i}\leq B$. 

\item \textbf{Outliers:} 
Some of the samples may be ``outliers''. 
We say that a point 
$(\va_o, \vb_o)$ is an outlier if, for all $j \in [M]$, $d_{\text{cluster}}(G_j, (\va_o, \vb_o))> \tau$. 
\end{enumerate}

\end{definition}

For \Cref{alg:EM} to converge, 
we will require good initial clustering (in Section~\ref{sec:experiments}, we show that using a simple Euclidean clustering 
or more modern alternatives, like SegmentAnything (SAM)~\cite{Kirillov23arxiv-SAM}, suffices). 
Here we formalize what it means to have good initial clustering: 
\begin{definition}[Good Clustering]\label{def:initial_conditions}
In the setting of Definition \ref{set:problem},
we say that the initial clustering 
$\cH := \{H_1, \dots, H_{K} \}$ with $K \geq M$ is $(\tau,\alpha, m_0)$-good, 
if it is a partition of the correspondences $S$ (as defined in Definition \ref{set:problem}) 
satisfying:
\begin{enumerate}
\item \textbf{$\tau$-connected:} 
For all $j \in [K]$,  $H_j$ is $\tau$-connected.
\label{item:tau-connected}
\item \textbf{Large Initial Clusters:} 
For all $j \in [K]$, $|H_j| \geq m_0$. \label{item:LargeObjects}
\item \textbf{Identifying cluster:} 
 For each ground truth cluster $G$, let $\cH_G := \{ H \in \cH \mid |H \cap G| > 0 \}$ and $H^* := \argmax_{H \in \cH_G} |H|$. Then, for some $\alpha > 1$, $|H^*| > \alpha \max_{H \in \cH_G \setminus \{H^*\} }|H|$.
\label{item:identifying-cluster}
\end{enumerate}
\end{definition}

Intuitively, the last condition captures the idea that for any ground truth cluster $G$, the largest cluster having a nonzero intersection with $G$, namely $H^*$, is notably larger than all other initial clusters having a nonzero intersection with $G$. 

\begin{theorem}[Expectation-Maximization Guarantee]
\label{thm:EM}
In the setting of Definition \ref{set:problem}, 
assume that the initial clustering $\cH$ is 
$(\tau,\alpha,m_0)$-good in the sense of Definition \ref{def:initial_conditions}, for some sufficiently large $m_0$.
Then, running \Cref{alg:EM}, with high probability (dependent on $m_0$),
returns $\cH'$, a partition of the set of correspondences $S$, {containing each of the ground truth clusters} (\ie the objects and background). 
\end{theorem}

\emph{Intuition:}
The proof of~\cref{thm:EM} follows by noting that the initial clustering results in a partition of each ground truth cluster such that one of the partitions is notably larger than the rest. 
As the algorithm progresses, the M-step assigns more points to the biggest estimated cluster until it exactly matches the ground truth object which contains it. 
This happens because the likelihood $W_{i,j;\tau}$ is maximized by the largest cluster, due to the 
presence of the weight $\pi_j$ (\ie the cluster size).

\emph{Proof Sketch:}
We make the following observations, which together imply that the final clusters produced by~\Cref{thm:EM}  include the ground truth clusters. 
First, {each ground truth-cluster is partitioned by the initial clustering}. This is because
the $\tau$-connected subsets of the data either consist of samples that are entirely contained in one of the $G_j$ or (possibly) the set of outliers. Since the initial clustering $\cH$ exclusively consists of $\tau$-connected subsets, 
each element of $\cH$ is either a subset of $G_j$ for some $j$ or entirely consists of outliers.

Now 
suppose the weights for the clusters are given by $\pi_1, \dots, \pi_{K}$.
Without loss of generality, suppose $\{ H_1, \dots, H_t \}$ form a partition of $G_1$ 
with $\pi_1 \geq \dots \geq \pi_{t}$. Since $\cH$ is $(\tau,\alpha, m_0)$-good,
we know that $\pi_1 > \alpha \pi_2$ because of \Cref{item:identifying-cluster} in Definition~\ref{def:initial_conditions}. 
Our second claim is that in each iteration of the $M$-step,
elements of $S$ that are $\tau$-close to the largest cluster $H_1$ are assigned to $H_1$. 
To see this, consider a point $(\va_i, \vb_i) \in G_1 \setminus H_1$ which is $\tau$-close to $H_{1}$ (if no such point exists, then $H_1 = G_1$). We show that for the point $(\va_i, \vb_i)$, the likelihood term $W_{i,j;\tau}$ is maximized when $j=1$. 

Since Horn's Method is consistent (\ie for a sufficiently large sample size, the algorithm converges to the true solution in the presence of zero-mean  noise) and the domain of the point-cloud is bounded by a ball of radius $B$, $h_j := \MR_j^{(r)}\va_i + \vt_j^{(r)}$ and $\sigma_j$ estimate $g_1$ and $\sigma$ up to an additive error of $\eps'$, according to standard concentration results. 
Choosing sufficiently large $m_{\min}$ and $m_0$ ensures that the sample size is large and consequently that $\eps'$ is small.
This is enforced by the algorithm, which deletes candidate clusters of size less than $m_{\min}$.
Now note that $\argmax_j W_{i,j;\tau} 
 = \argmax_j \pi_j \phi_j(\vb_i \mid \va_i) 
 = \argmax_j \pi_j \exp(-\norm{\vb_i - h_j(\va_i) }^2/{\sigma_j}^2) {/\sigma_j}$. 
Since $\eps'$ is sufficiently small for large sample size, we get
$W_{i,j;\tau}= \pi_j (1\pm \eps) \exp(-\norm{\vb_i - g_1(\va_i)}^2/{\sigma}^2)/\sigma$, where $\eps$ is a function of $\eps', \sigma$, and $B$.
%
%
%
Since $\exp(-\norm{\vb_i - g_1(\va_i)}^2/{\sigma}^2)/\sigma$ is a constant with respect to $j$, it does not affect the maximization. 
This implies 
$\argmax_j W_{i,j;\tau}$ is essentially determined by $\argmax_j \pi_j (1\pm \eps)$. Choosing small $\eps$ to ensure $(1+\eps)/(1-\eps) < \alpha$ and recalling that $\pi_1 > \alpha \pi_2$, we see $\argmax_j W_{i,j;\tau} = \argmax_j \pi_j (1\pm \eps) = 1$. Since $\pi_1$ keeps increasing in size at each iteration, eventually all the elements of $G_1$ will collect into $H_1$. 
\qed



\begin{remark}[Novelty]
Similar to theoretical analyses of the EM algorithm in prior work in the context of learning a mixture of linear regressions, we require a good initialization, see,~\eg \cite{kwon20aistats-EM-MLR}. However, 
contrary to related work, we do not assume the regressors (roughly speaking, the vectors $\va_i$) to follow a Gaussian distribution, which would be too strict of an assumption in practical multi-modal registration problems. 
\end{remark}

\section{Experiments}

\label{sec:experiments}
We conduct a wide range of experiments on both synthetic and real-world datasets. The synthetic datasets are  PASCAL3D+~\cite{Xiang14-Pascal3DPlus} and FlyingThings3D~\cite{Mayer16CVPR-FT3D} and the real-world dataset is from 
KITTI~\cite{Geiger13IJRR-KITTI}. 
We test the proposed approach against Sequential RANSAC~\cite{Torr98royalp-seqransac} and T-Linkage~\cite{Magri14CVPR-tlinkage}, and show it
dominates these baselines on the multi-model 3D registration problem, and its performance is further improved by using SegmentAnything (SAM) \cite{Kirillov23arxiv-SAM} as initialization.

\subsection{Baselines and Initialization}

We compare our approach against T-Linkage~\cite{Magri14CVPR-tlinkage} and Sequential RANSAC (SRANSAC)~\cite{Torr98royalp-seqransac}. 
We also include a Naive baseline that applies Horn's method~\cite{Horn87josa} to compute a pose estimate for each initial cluster.

\myParagraph{T-Linkage~\cite{Magri14CVPR-tlinkage}} 
T-Linkage computes the distance between pairs of clusters and iteratively merges the closest pair of clusters. 
For each cluster $j$ (with associated pose $(\MR_j,\vt_j)$), T-Linkage defines a \emph{preference function} for point $i$ as
\begin{equation}
\psi_{j, i} = \e^{- d_{j, i}/\tau_t} \text{ if } d_{j, i} \leq 5 \tau_t \text{ or } 0 \text{ otherwise},
\label{eq:preffunc}
\end{equation}
where $d_{j, i} := \|\vb_i \!-\! \MR_j \va_i \!-\! \vt_j\|$.
{It then uses the preference functions to compute distances between pairs of clusters, and terminates when the distance between every pair is large.
}

\myParagraph{Sequential RANSAC~\cite{Torr98royalp-seqransac}} This baseline sequentially applies RANSAC and tries to recover one object at a time. After each RANSAC execution, the correspondences selected as inliers by RANSAC are used to compute a pose estimate and then removed to facilitate the search for other objects. 

\myParagraph{Initialization} SRANSAC does not require an initial guess for the clusters, while 
T-Linkage and our approach do.
In PASCAL3D+ we use Euclidean clustering on $\{\va_i\}_{i=1}^n$ to obtain initial clusters for both T-Linkage and EM.
In KITTI and FlyingThing3D, we ablate the effect of initial clustering, by initializing T-Linkage \cite{Magri14CVPR-tlinkage} and our method with SegmentAnything (SAM)~\cite{Kirillov23arxiv-SAM} or Euclidean clustering. We tune both initialization approaches to generate around 100 clusters.


\subsection{Metrics}
We use three main performance metrics for evaluation. 

\myParagraph{Per-Point Error ($\downarrow$)} 
This metric evaluates the average mismatch between the ground-truth and estimated point clouds associated with each object (including the background).
It first segments the first point cloud according to the ground-truth clusters 
to obtain $\va\at{i}, i=1,\ldots,M$, and then applies the ground-truth transformation to each object to obtain $\vb\at{i}, i=1,\ldots,M$. It repeats the same process using the estimated clusters and transforms to compute $\hat{\vb}\at{j}, j=1,\ldots,K$. 
Then, each estimated object $i$ is associated with the ground truth object $j$ that has the largest intersection with $i$ (\ie the largest number of points in common), 
and the Chamfer distance between $i$ and $j$ is recorded. The point error is defined as the average Chamfer distance across objects. 

\myParagraph{Rotation and Translation Error ($\downarrow$)} This metric evaluates the distance between the estimated and ground-truth poses. For each estimated object $H_j$, we find every ground-truth object $G_k$ that has a non-zero intersection with $H_j$. 
%
Then, we compute the \textit{translation error} as the Euclidean distance between the estimated and ground-truth object translation; 
the 
 \textit{rotation error} is the angular distance~\cite{Hartley13ijcv} between the estimated  rotation and the ground truth rotation.
 {The final translation and rotation errors of each estimated cluster are the weighted averages of the errors, where the weights are computed as ${|H_j \cap G_k| } / {|H_j|}$.}

 \myParagraph{Intersection over Union ($\uparrow$)} This metric evaluates the quality of the estimated clusters by assigning a ground-truth cluster to each estimated cluster with the largest intersection and calculating the average Intersection over Union (IoU). 



\begin{table}[t]
\caption{Results for synthetic PASCAL3D+ dataset. Experiment 1: Noiseless. Experiment 2: with additive Gaussian noise. Experiment 3: with additive Gaussian noise and 2 objects with the same motion.}
\resizebox{\columnwidth}{!}{
\begin{tabular}{l|l|l|l|l|l|l}
\hline
                    & Metric & \multicolumn{5}{c}{Method}    \\
\cline{3-7}
                    & (Mean) & Naive & SRANSAC & T-Linkage   & Ours (Vanilla) & Ours \\
\hline
\hline
\multirow{4}{*}{1}  & Per-point Error [m]    &   0.0454 & \textbf{9.56e-15} &  0.163 &  \textbf{9.56e-15} & \textbf{9.56e-15}    \\
                    & Rotation error [deg]     &   46.8 & \textbf{8.69e-7} & 19.6 &  \textbf{8.69e-7} &  \textbf{8.69e-7}    \\
                    & Translation error [m]  &   0.487 & \textbf{3.81e-15} & 0.242 & \textbf{3.81e-15} & \textbf{3.81e-15}    \\
                    & IoU                &   0.698 & \textbf{1.0} & 0.723 & \textbf{1.0}  & \textbf{1.0}     \\
\hline
\multirow{4}{*}{2}  & Per-point Error [m]   &   0.0550 & 0.332 & 0.199 & 0.0135 &  \textbf{0.00516}     \\
                    & Rotation error [deg]    &   81.6 & 74.2 &  27.6 & 2.42 & \textbf{1.53}    \\
                    & Translation error [m]  &   0.858  & 0.796 & 0.287 & 0.0286 & \textbf{0.0165}   \\
                    & IoU                &    0.668 & 0.450 & 0.785 & 0.908 & \textbf{0.964}    \\ 
\hline
\multirow{4}{*}{3}  & Per-point Error [m]    & 0.0104 & 0.321 & 0.215 &  0.00860 & \textbf{0.00776}   \\
                    & Rotation error [deg]    & 75.3 & 67.3 & 15.9 & 1.48 & \textbf{1.12}     \\
                    & Translation error [m]  & 5.83 & 1.33 & 0.300 &  \textbf{0.0200} &  0.0499       \\
                    & IoU                & 0.755 & 0.398 & 0.729 & 0.825 & \textbf{0.970}      \\                                                               
\hline
\end{tabular}
}
\label{table:pascal3d}
\vspace{-6mm}
\end{table}

\begin{figure*}[t]
    \centering
    \begin{subfigure}{0.48\textwidth}
        \centering
        \includegraphics[width=\linewidth]{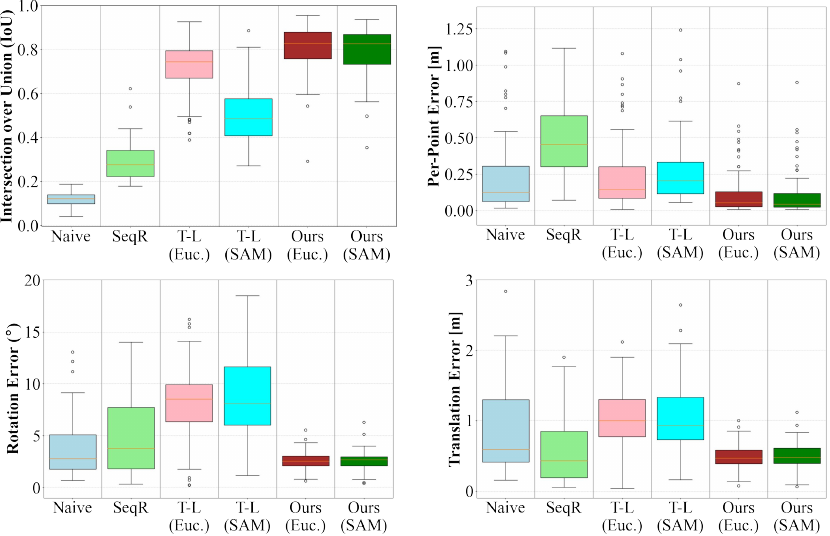}
        \caption{FlyingThings3D Results}
        \label{fig:sub1}
    \end{subfigure}
    \hfill
    \begin{subfigure}{0.48\textwidth}
        \centering
        \includegraphics[width=\linewidth]{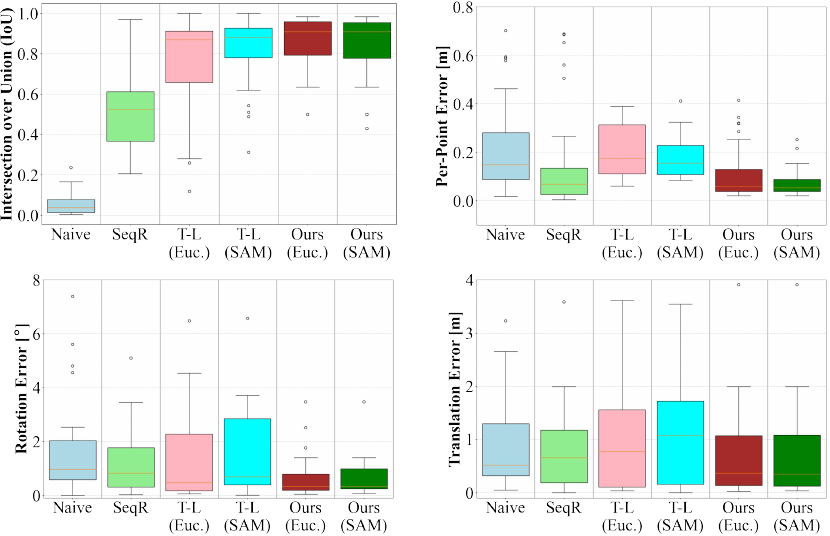}
        \caption{KITTI Results}
        \label{fig:sub2}
    \end{subfigure}
    \caption{Results (IoU, per-point error, rotation error, and translation error) on (a) FlyingThings3D and (b) KITTI. We evaluate two variations of our method with two different initializations (SAM and Euclidean) and four baselines: Naive, Sequential RANSAC (SeqR), T-Linkage with SAM initialization (T-L (SAM)), and T-Linkage with Euclidean initialization (T-L (Euc.)).}
    \label{fig:boxplots-main}
\vspace{-6mm}
\end{figure*}

\subsection{PASCAL3D+}
\myParagraph{Experimental Setup}
PASCAL3D+ \cite{Xiang14-Pascal3DPlus} is a synthetic dataset for 3D object understanding. We choose 7 objects from the dataset and downsample the vertices of their CAD models to form object point clouds (22,395 points overall). To generate point cloud pairs $\{(\va_i, \vb_i)\}_{i=1}^{n}$, we randomly sample 7 transformation matrices and apply them to each object point cloud. In this dataset, there are no outliers, and we want to test the capability of the compared techniques to tell the 7 objects apart and estimate their motion (Experiment 1).
We also repeat the same test in two more challenging settings, where we 
add zero-mean {Gaussian noise} with standard deviation $0.03$m to the point cloud $\{(\vb_i)\}_{i=1}^{n}$ (Experiment 2), and where ---in addition to the noise--- we 
assume 2 of the 7 objects have the same motion and we still want to tell them apart (Experiment 3).
We test our method with and without the distance term in{~\eqref{eq:likelihood}}; we denote the latter as ``Ours (Vanilla)''.
We use Euclidean clustering (with 100 clusters) to obtain the initial clusters for Naive, T-Linkage, and our methods.
For our method, we set $\tau = 1.5$m, $m_\min = 4$, $T = 10$.
We set $\tau_t = 0.2$m in T-Linkage. 
For SRANSAC, we use 0.01m as the inlier threshold for experiment 1 (noiseless case) and 0.5m for the others and use 1000 max iterations.
Results are averaged over 100 runs.

\myParagraph{Results} 
Table~\ref{table:pascal3d} shows the results obtained with the compared techniques in the three settings described above. 
In the noiseless case, SRANSAC and our methods achieve perfect scores (\ie errors are numerically zero). 
However, in the noisy experiments, our method outperforms other methods across metrics. This is because our method automatically adjusts the noise variance at each iteration and computes the likelihood function accordingly.
In Experiment 3, where we enforce two objects' motion to be identical, although both variants of our methods still estimate the poses with similar errors, the one with the distance term (``ours'') stands out in terms of IoU: with the distance term, our method can identify two objects that are relatively far apart even if they have the same motion. 
Naive reflects the quality of the initial clustering and typically leads to poor IoU. 
T-Linkage consistently improves over the initial clustering (Naive) in all experiments, but is not competitive with SRANSAC and our methods.
The runtime for SRANSAC is 30ms, while T-Linkage takes about 2s. Our method without the distance term takes about 800ms in Python on a Macbook Pro with M1 Pro chip. Adding the distance term in our method increases the runtime to a couple of seconds, since we have not optimized the distance computation.

\subsection{FlyingThings3D}\label{sec:flying}
\myParagraph{Experimental Setup}
The FlyingThings3D \cite{Mayer16CVPR-FT3D} dataset has randomly moving objects from ShapeNet~\cite{Wu15cvpr-shapenets}. The dataset provides RGB images, segmentation masks, depth maps, and disparity maps. We construct the point clouds with ground-truth scene flows by back-projecting pixels to 3D points using disparity maps. 
In our method, to construct correspondences, we use the state-of-the-art scene flow model CamliRAFT~\cite{Liu22CVPR-camliflow} on RGB images and downsampled point cloud $\va$ with 32,768 points to get predicted scene flow $\vf$ and add predicted scene flow on the point cloud to get the next frame's point cloud $(\vb=\va + \vf)$. 
Using the object-level segmentation masks, we obtain ground-truth poses for each object cluster by running Horn's method on each object's point cloud and its counterpart displaced by the ground-truth scene flow.

We set the distance threshold in our method to be 5m, since the diameters of most objects are about 3m in the FlyingThings3D dataset. Then, for T-Linkage we set the constant $\tau_t$ to be 1m in \eqref{eq:preffunc}. For SRANSAC, we use 0.2 as the residual threshold and 1000 (maximum) iterations.

\myParagraph{Results}
We show the comparison between our method and other baselines on FlyingThings3D in Fig. \ref{fig:sub1}. As shown in the boxplots, our method outperforms other baselines. In particular, our method recovers the object clusters consistently (\ie it has the highest IoU) and works well with different kinds of initialization methods across all metrics. SRANSAC achieves a very low IoU score because it recovers over 50 clusters which is a lot more than the ground truth (about 10 clusters). T-Linkage clusters the point cloud better using Euclidean clustering which is considered as a weaker initialization method than SAM. This is because, in the FlyingThings3D dataset, since the objects are relatively far apart, the initial clusters from Euclidean clustering are significantly better than SAM. For our method, IoU scores are almost the same. This shows that our method is less sensitive to the quality of initial clustering. 


\subsection{KITTI}
\myParagraph{Experimental Setup}
The KITTI {\cite{Menze15cvpr-KITTIsceneflow,Geiger13IJRR-KITTI}} scene flow dataset consists of 400 {scenes} split into training, validation, and testing datasets with RGB images and depth. We only use the validation set because no ground-truth optical flow is provided in the testing set for us to evaluate. 

To run our method, we follow the same steps done for the FlyingThings3D dataset to construct correspondences and obtain initial clustering. 
Since KITTI only provides semantic masks, to compute ground truth poses and (instance-level) clusters we perform Euclidean clustering only on the point clouds with the car label in the ground-truth semantic segmentation and merge everything else as background, resulting in a masked point cloud with a cluster for each car and another cluster for the background. Then, we calculate the pose for each cluster, similar to FlyingThings3D. We use the same parameter choices as in Section~\ref{sec:flying} since the main moving objects here are cars which are also about 3m in length.

\myParagraph{Results}
In Fig. \ref{fig:sub2}, we compare our method and other baselines on the KITTI dataset. Our method outperforms other baselines. SRANSAC still suffers from over-segmenting as in the FlyingThings3D experiment. Since SAM performs better on KITTI (compared to Euclidean clustering), 
T-Linkage exhibits slightly better performance with the SAM initialization.

\section{Conclusion}
\label{sec:conclusion}
We investigated a variation of the 3D registration problem, named \emph{multi-model 3D registration}, that simultaneously recovers the motion of multiple objects in point clouds. We proposed a simple approach based on Expectation-Maximization (EM) and established theoretical conditions under which the EM scheme recovers the ground truth. We evaluated the EM scheme in both synthetic and real-world datasets ranging from table-top scenes to large self-driving scenarios and demonstrated its effectiveness. 

\bibliographystyle{IEEEtran}
\bibliography{references/refs}


\end{document}